\documentclass[conference]{article}

\usepackage[utf8]{inputenc}
\usepackage{comment}
\usepackage{macros}
\usepackage{xspace}
\usepackage{xcolor,colortbl}
\usepackage{hyperref}
\usepackage[misc]{ifsym}

\newcommand{\orc}{\includegraphics[height=\fontcharht\font`A]{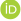}}

\newcommand{\sump}{SUTP\xspace}

\title{A Constraint Programming Model for Scheduling the Unloading of Trains in Ports: Extended}

\author{Guillaume Perez\href{mailto:guillaume.perez06@gmail.com}{\Letter}
\href{https://orcid.org/0000-0001-6473-583X}{\orc}
Ga\"{e}l Glorian\href{mailto:gael.glorian@huawei.com}{\Letter}
\href{https://orcid.org/0000-0002-0843-5987}{\orc},\\
Wijnand Suijlen\href{mailto:wijnand.suijlen@huawei.com}{\Letter}
\href{https://orcid.org/0000-0001-6450-5620}{\orc},
Arnaud Lallouet\href{mailto:arnaud.lallouet@huawei.com}{\Letter}
\href{https://orcid.org/0000-0002-4318-356X}{\orc}\\
\textit{Huawei Technologies Ltd,
Boulogne-Billancourt, France}
}
\date{}

\begin{document}
\maketitle

\begin{abstract}
In this paper, we propose a model to schedule the next 24 hours of operations in a bulk cargo port to unload bulk cargo trains onto stockpiles. 
It is a problem that includes multiple parts such as splitting long trains into shorter ones and the routing of bulk material through a configurable network of conveyors to the stockpiles.
Managing such trains (up to three kilometers long) also requires specialized equipment.
The real world nature of the problem specification implies the necessity to manage heterogeneous data.
Indeed, when new equipment is added (e.g. dumpers) or a new type of wagon comes in use, older or different equipment will still be in use as well.
All these details need to be accounted for. In fact, avoiding a full deadlock of the facility after a new but ineffective schedule is produced.
In this paper, we provide a detailed presentation of this real world problem and its associated data.
This allows us to propose an effective constraint programming model to solve this problem.
We also discuss the model design and the different implementations of the propagators that we used in practice.
Finally, we show how this model, coupled with a large neighborhood search, was able to find 24 hour schedules efficiently.
\end{abstract}

\section{Introduction}
The logistics of bulk cargo ports involves a lot of activities that can benefit from optimization-based solutions.  
Port scheduling in particular is a prolific research area \cite{goel2015constraint,kizilay2020constraint,sacramento2020constraint,he2021modeling,jimenez2023constraint} with problems like container packing, crane scheduling, or berth planning.
The logistics of bulk cargo includes the transport from the mine to a port, from a port to another port, and then from the port to the client.
These bulk materials are usually not packed into containers, but rather handled in bulk (e.g. wagons, silos, stacks) and sold by the ton.
Such problems are different from classical containers scheduling problems as bulk material can be arbitrarily split and is quite similar to a continuous resource.

Scheduling is one of the most successful applications of constraint programming (CP) \cite{baptiste2001constraint,laborie2018ibm}.
From metal factory \cite{gay2014continuous}, hospital \cite{cheng1997nurse,hashemi2016constraint} to music composition \cite{roy2016enforcing}, constraint programming models provide an efficient solution to scheduling problems.

This paper presents a constraint programming model for Scheduling the Unloading of Trains in Ports (\sump) used in an industrial context.  
This is a scheduling problem for unloading bulk cargo trains in a bulk port.
It is the first sub-problem of a larger logistics chain from the mine to the client via bulk ports where the cargo from  bulk cargo trains is unloaded onto stockpiles.
The goal is not to describe a flexible commercial software but more to show how efficient techniques used in Constraint Programming can be applied to a real application and lead to immediate success.
Several instances of coal scheduling problems can be found in the litterature \cite{DBLP:journals/heuristics/BelovBSS20,DBLP:journals/candie/PaulaBEMS19,DBLP:journals/scheduling/AngelelliKKS16}, including the complementary problem of vellel loading \cite{DBLP:conf/cpaior/BelovBSS14} that takes place immediately after the problem presented here.

\begin{figure}[ht]
    \centering
    \includegraphics[width=11cm]{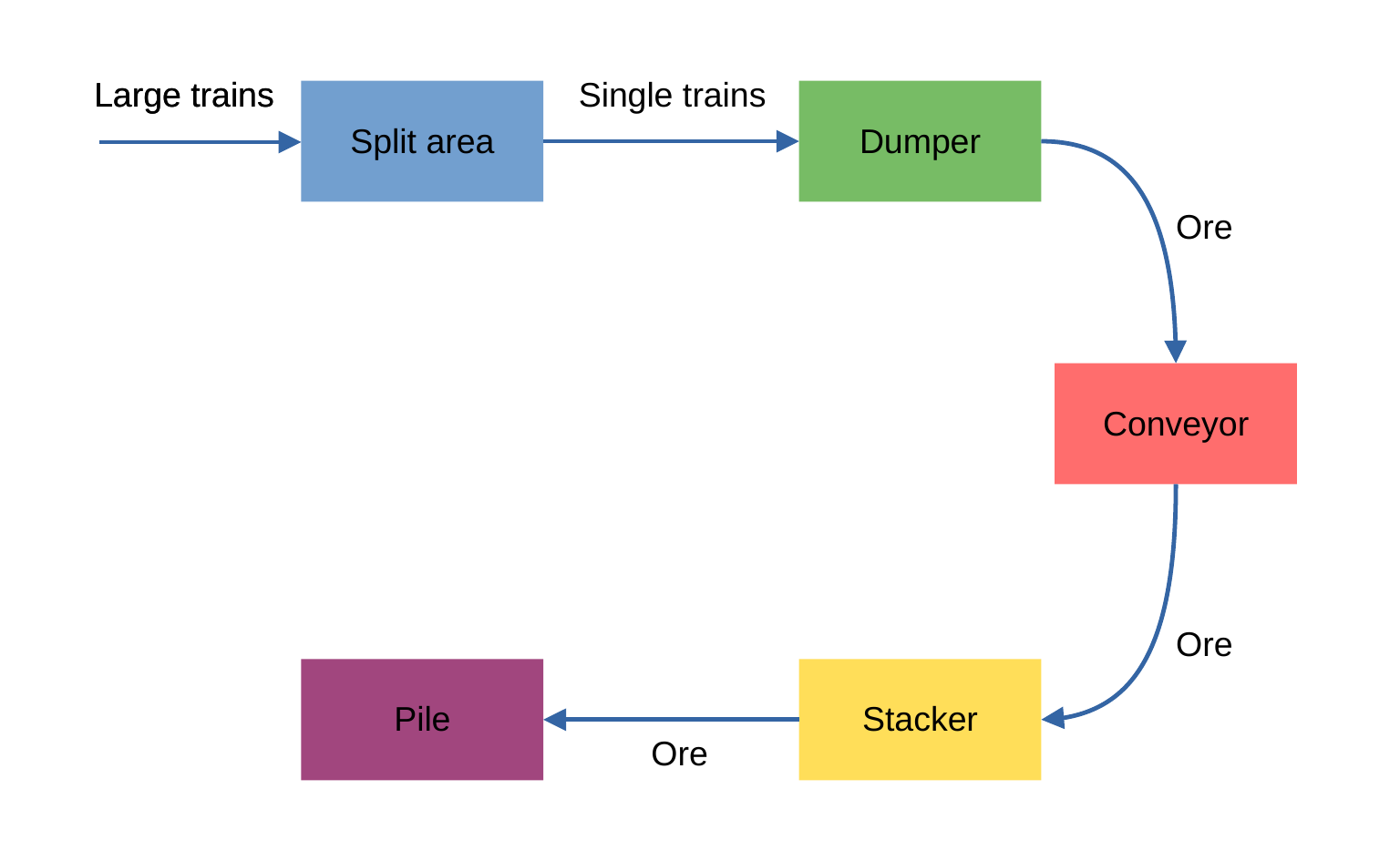}
    \caption{Operations workflow of the port}
    \label{fig:workflow}
\end{figure}

Section \ref{sec:ProblemDef} gives a global description of the problem to be solved.
This problem is a complex system including multi-stage decisions.  Its workflow, depicted in Figure \ref{fig:workflow}, includes five different stages. In the first stage, all \emph{big trains} arrive on a shunting yard. The second stage splits the big trains into smaller ones so-called \textit{unit trains}. In the third stage, dumpers unload the unit trains  onto conveyors. At the fourth stage, the conveyors transfer the cargo to the stockpile area. It is important to note that the network of conveyors is configurable to establish a path from the dumper to the stacker. At the fifth and last stage, the cargo is collected by a stacker, which is a large mobile machine that piles bulk material onto a stockpile from the top. Stockpiles are organized in rows in front of the boat berthing zone. A particular stacker can only access stockpiles from a single row. Indeed, stackers move on rail only in a straight line.

Section \ref{sec:dataAnalysis} further analyses the common characteristics of instances (i.e. the port configuration itself).
Indeed, exploiting the port topology allows us to define an ad hoc model to efficiently solve this problem.
Then, section \ref{sec:model} proposes a constraint programming model.
This model is a dedicated model made for the port described in section \ref{sec:dataAnalysis}.
We propose several variants of the model, which we profiled in order to select the most efficient one.
Finally, the experimental section shows how a large neighborhood search on the constraint model is able to efficiently solve industrial size instances using our internal CP solver.

\section{Scheduling the Unloading of Trains in Ports (\sump)
} 
\label{sec:ProblemDef}

The \sump problem is a supply-chain problem where given a port topology
and an arrival schedule of  big trains (big trains are up to three kilometers long), 
a complete schedule for the various bulk material handling equipment must be found.
The arrival schedule of big trains is updated every 15 minutes and each time a new feasible schedule must be found for the next 24 hours.
In this section, we first present high level definitions for the different parts of the problem.
Then, we provide the complete list of constraints that must be satisfied by the schedule.

The starting point is the arrival of a \textit{big train}, which is a set of 1 to 4 subsets of wagons that we call \textit{unit trains}.
\textit{Unit trains} represent the smallest unit that can be processed by the first type of bulk material handling equipment, the \textit{dumper}.  

The first step of the problem is to decide which splitting plan must 
be applied to the \textit{big train} to extract the \textit{groups} of \textit{unit trains}.
This part is called the \textit{splitting plan}. 
A splitting plan must follow a set of rules and has an impact on the makespan.
Note that splitting is the only action that does not require any special equipment.

Subsequently, once the \textit{groups} are defined, they must be routed to the \textit{dumpers}.
\textit{Dumpers} are machines that unload the wagons onto \textit{conveyors}.
Dumpers can only accept some types of trains, some types of bulk material, etc.
Next, the conveyor graph represents paths that link the \textit{dumpers} to the \textit{stackers}.
While dumpers are used to unload trains, stackers are machines that temporarily stockpile the bulk material.
The tuple \textit{(dumper, conveyors, stacker)} is called the \textit{equipment flow} of a train.
Each train must be scheduled using an equipment flow.
Finally, \textit{stackers} can usually reach only one or two of the \textit{stockpiles}.
The stockpile area is laid out in a grid, where the stackers can move along a row over rail. 
Note also that the stockpiles already have some type of bulk material
and have a maximum capacity.
While there are a large number of stockpiles, usually only a small number can accept cargo from a given train.
In summary, the steps for each big train are:
\begin{enumerate}
    \item Split the big train into groups of unit trains.
    \item Bring the unit trains to the dumpers.
    \item Dump the cargo from the unit trains using matching dumpers.
    \item Transport the cargo via connected conveyors from the dumpers to the stackers.
    \item Pile the cargo on valid stockpiles using the available stackers.
\end{enumerate}

\subsection{Notations and Definitions}
We provide here a list of keywords that we will use throughout this paper. 
A \textit{big train} is a large train composed of many wagons, usually hundreds. 
It is composed of \textit{unit trains}.
We call \textit{length} the number of unit trains in a big train.
A \textit{unit train} is a smaller train composed of several identical wagons (same bulk material, same type of wagon).
This is the minimal unit that can be scheduled on a dumper. 
The objective of the \sump problem is to schedule all the unit trains.
A \textit{dumper} is a machine that unloads the cargo from a unit train onto a conveyor.
In practice, it is located in a huge building inside of which it processes the wagons one-by-one.
A \textit{conveyor} is a machine that transports bulk material between two points.
A \textit{stacker} is a machine that receives bulk material from a conveyor and piles it onto a \textit{stockpile}.
It can move between multiple stockpiles while accepting bulk material from the same conveyor.
A \textit{stockpile} is a pile of bulk material that can grow to a certain capacity.
A \textit{group} is a set of unit trains that can be processed by the same dumper (i.e. a dumper has enough capacity to accept a group of 2 or more unit trains).
An \textit{equipment flow} is a set of bulk material handling systems that is used to transport the bulk material from dumpers to stackers.
It is a \textit{path} in the equipment graph.

The specification of each \sump problem instance is composed of two parts: a static part and a dynamic part.
The static part is the specification of the port and the rail network.
This part remains the same all the time. The only change that may occur is that some sector of the port may be unavailable.
Otherwise, the global topology stays the same.

\begin{figure}
    \centering
    \includegraphics[width=10cm]{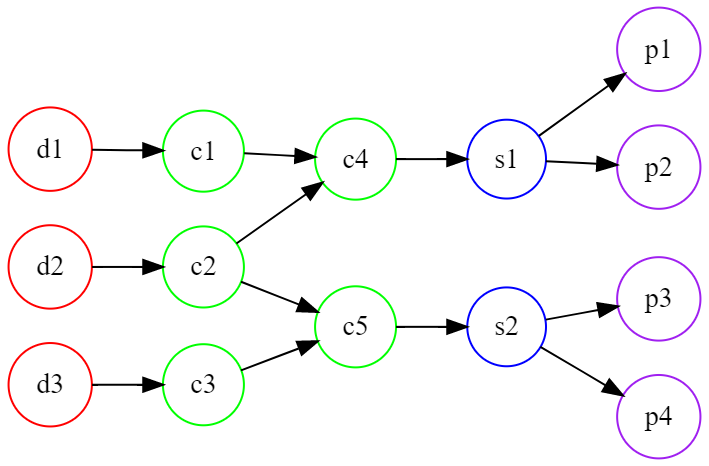}
    \caption{Example of equipment graph.}
    \label{fig:equipmentFlow}
\end{figure}

\paragraph*{Static data}
The port is composed of a large equipment flow graph $G$.
Dumpers ($D$) are linked to conveyors.
Conveyors ($C$) are linked to other conveyors and to stackers.
Stackers ($STK$) are then linked to stockpiles ($P$).
Figure~\ref{fig:equipmentFlow} shows an example equipment graph.
In this example, there are three dumpers ($d_i$, red nodes), which are linked to conveyors ($c_i$, green nodes).
Moreover, the conveyors are linked together and to the stackers ($s_i$, blue nodes).
Finally, the stackers are linked to the stockpiles ($p_i$, purple nodes).
Any path from a dumper to a stockpile is an equipment flow (e.g. $d_2, c_2, c_4, s_1, p_1$).

The rail network between the shunting yard and dumpers is also always the same. 
This implies that the splitting time and the transportation time from the shunting yard to the dumpers is constant.
Additionally, the clearance time for bulk material handling equipment and the like is also constant.
The only varying part of the specification from one instance to another is the arrival schedule of big trains.

\paragraph*{Dynamic data}
The goal of each problem instance is to schedule a set of big trains.
Each big train $b \in B$ is specified by an arrival date and time ($\mathtt{arrival}_b$) and by its composition of unit trains.
The specification of each unit train $t \in T$ consists of elements, such as its height ($height_t$), its load ($\mathtt{load}_t$), its type of cargo ($CType_t$), and its type ($type_t$).

\paragraph*{Objectives}
The main objective of this problem is to minimize the completion time of the schedule.
In practice, depending on the client's needs, more objectives can be added, such as maximizing the throughput during the next few hours or even minimizing the waiting time of unit trains.
For simplicity's sake, these objectives are not considered in this document.

\subsection{Constraints}
Here, we summarize the different constraints that the model must satisfy.
This list is complete and will be reused later to map the constraints of the CP model
and the constraints of the problem.

\begin{constraint} \label{ctr:equipeSched}
\textit{Any piece of equipment can service at most one train at a time.}
This is the main scheduling constraint, 
that concerns all equipment (dumpers, conveyors, stackers, and stockpiles).
Indeed, two unit trains cannot be scheduled on the same dumper at the same time.
This is also true for any equipment as in most scheduling problem.
In addition, once a dumper has started processing a group of unit trains, it cannot be stopped.
This first constraint is the core scheduling constraint of the problem and
will be the one driving the CP model definition.
\end{constraint}

\begin{constraint}\label{ctr:splitPlan}
\textit{Incoming big trains must follow the splitting rules.}
Let us consider a big train with length K (i.e. containing K unit trains). We can represent its composition by the string \textit{ABC...K}, where each letter represents a unit train.
In the current set of instances, the maximum length is 4 (i.e. a big train is composed of at most of 4 unit trains).
Hence, the composition of the largest big train can be represented by the string \textit{ABCD}.
The splitting rule states that a group must be a substring of the composition string.
For example, \textit{groups} AB and CD can be extracted from the big train, but AD cannot.
Note that this rule does not impose in which order groups have to be processed.
\end{constraint}

\begin{constraint}\label{ctr:splitTime}
\textit{Each split operation has a duration.}
It usually is around 30 minutes in our problem instances.
For example, consider a big train ABCD. 
Creating groups AB, C, and  D will cost 2 times 30 minutes.
Note that the complete splitting plan must be finished
before any group can be processed by the next stage.
\end{constraint}

\begin{constraint}\label{ctr:DumperTransport}
\textit{Moving a group of unit trains from the shunting yard to a dumper has a duration.}
It usually is around 90 minutes in our problem instances.
This duration is always the same because it only depends on the rail network of the port.
\end{constraint}

\begin{constraint}\label{ctr:groupDumper}
\textit{Each unit train in a group must use the same dumper, but can use different equipment flows.}
This constraint implies that two unit trains in a group can have different equipment flows, but they must share their dumper.
\end{constraint}

\begin{constraint}\label{ctr:groupOrder}
\textit{The unit trains from the same group are processed sequentially and in order.}
The order is determined by the big train composition. 
In contrast to different groups from a big train, that can be processed in any order,
the unit trains inside the same group must remain in that order during unloading.
For example, for a given big train ABCD split in half in AB-CD. The two groups AB and CD can be processed in any order (AB then CD, or CD then AB), but inside the group AB (resp. CD), A (resp. C) will always be processed before B (resp. D).
\end{constraint}

\begin{constraint}\label{ctr:DumperClean}
\textit{A dumper requires a clearance time after each group.}
Trains take time to clear a dumper and time depends on the dumper specifications.
Indeed, for each type (and especially size)
of dumper, the clearance time will be different.
In our problem instances, we have two different clearance times.
\end{constraint}

\begin{constraint}\label{ctr:DumperCtr}
\textit{A dumper can only process groups that match its length, cargo type, train height and train type.}
These four constraints are the configuration constraints of the dumpers. 
As we will show in the data section, for some restrictions, these constraints are loose, 
while for others, the constraints are highly restrictive.
\end{constraint}

\begin{constraint}\label{ctr:EquipmentFlow}
\textit{An Equipment flow is feasible if it forms a path from a dumper to a stacker in the equipment graph.}
This constraint simply imposes that the solution must be feasible. 
This implies that the equipment used by a train must be connected in the equipment graph.
\end{constraint}

\begin{constraint}\label{ctr:StackQTY}
\textit{Stockpiles are assigned to a certain type of bulk material and already store some quantity of it.} 
The stockpile can be chosen to store cargo only if the type matches.
The amount of unloaded cargo should not exceed the maximum capacity of the stockpile ($\mathtt{capacity}_p$). 
This constraint is a capacity  and configuration constraint.
For a train containing a given quantity and type of cargo, only a subset of stockpiles will be available.
In addition, these stockpiles have different capacity and a wrong choice in the selection of the first unit train may 
induce a bottleneck in the equipment graph for the next trains.
\end{constraint}

\begin{constraint}\label{ctr:Efficiency}
\textit{The processing time of a unit train depends on the efficiency of the equipment flow (ton/h) for the given train type.}
In practice, it is known that the throughput of a system is limited by the throughput of the slowest part.
In our case, the throughput is always limited by the dumpers.
\end{constraint}

\begin{constraint}\label{ctr:stkPiles}
\textit{Stackers can reach only a given subset of the stockpiles. }
This constraint comes from the fact that stackers usually are vertical conveyor belts sliding on a rail.
Just like quay cranes, they move slowly and cannot cross each other. So, most lanes only have a few stackers (in practice 1 or 2).
\end{constraint}

\section{Specifications} \label{sec:dataAnalysis}
The main problem instances are from an existing bulk port.
This section provides more insight on the specification of this port, particularly regarding size and topology.
Modelers should become familiar with it before making their models.

\subsection{Dumpers}
There are 13 dumpers in this port, named $CD1$ up to $CD13$.
$CD1$ to $CD9$ are older than $CD10$ to $CD13$.
This difference is the main reason why most configuration constraints exist.
Each dumper is linked to different conveyors and is defined by a particular configuration.
The following paragraphs detail the most important ones.

First, the train type is one of the characteristics of a unit train.
In this port, there are 3 possible train types, namely C80, C70, and C64.
For example, the train type C80 means that each wagon contains around 80 tons of bulk cargo,
C64 that each wagon contains around 64 tons of cargo, etc.
Nevertheless, a unit train typically carries around 5000 tons of cargo
and, thus, is typically composed of 54 to 62 wagons.

\begin{figure}
    \centering
    \includegraphics[width=6cm]{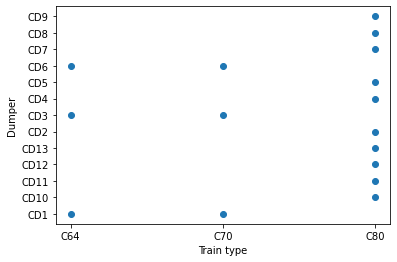}
    \includegraphics[width=6cm]{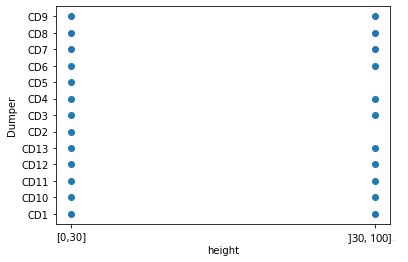}
    \caption{(left) Dumper/train type compatibility. A blue dot means that the dumper is compatible with the train type. 
    (right) Dumper/train height compatibility. A blue dot means that the dumper is compatible with the train height.}
    \label{fig:dumperTtype}
\end{figure}

A dumper is piece of equipment that unloads unit trains onto conveyors.
A unit train and a dumper are compatible if the dumper can work with its train type and height.
There are two main classes of heights: [0,30] and ]30,100].
These intervals represent the possible additional heights of the train with respect to the top of the wagon.
This information is enough to get the complete compatibility map.
Figure~\ref{fig:dumperTtype} (left) shows the compatibility between dumpers and train types.
As we can see, types C64 and C70 are compatible with only 3 dumpers
and there is no overlap with the ones that service C80 trains. 
This particular point will be important for the model definition.
Figure~\ref{fig:dumperTtype} (right) shows the compatibility between dumpers and train height.
As we can see, only two dumpers cannot process unit trains within the interval ]30,100].

\begin{figure}[t]
    \centering
    \includegraphics[width=6cm]{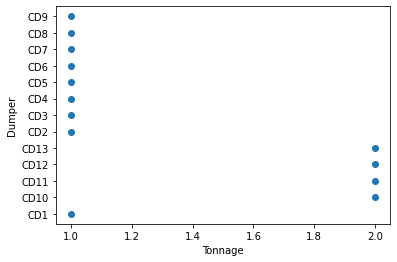}
    \includegraphics[width=6cm]{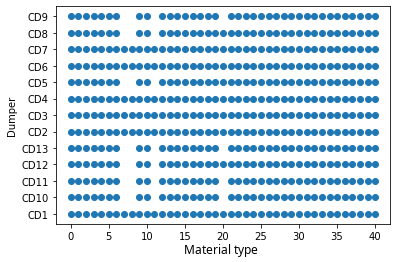}
    \caption{(left) Dumper/Length compatibility. A blue dot means that the dumper can receive a group of this length.
    (right) Dumper/type compatibility. A blue dot means that the dumper can receive a unit train with the given cargo type.}
    \label{fig:dumperTonnage}
\end{figure}

\textbf{Length}
Dumpers are made to work with a given length.
In our data, dumpers work with groups of size 1 or 2.
Recall that a set of unit trains that are processed by the same dumper together, is called a group.
Figure \ref{fig:dumperTonnage} shows the compatibility between dumpers and group length.
As we can see, $CD1$ to $CD9$ can service only one unit train at the time, while $CD10$ to $CD13$ can service only groups of exactly two unit trains.
Note that one of the reasons is that these are more recent and more efficient (in term of ton/h).
The principal difference is that dumpers $CD10$ to $CD13$ do not 
require any clearance time between two trains inside the same group.

\textbf{Cargo types}
There are many possible types of cargo. In our problem instances 42 different types occur. 
The dumpers accept most types but not everything.
Figure \ref{fig:dumperTonnage} shows the compatibility table.

\textbf{Clearance time}
Clearance time depends on the dumper type.
Trains take 60 minutes to clear dumpers $CD1$ to $CD9$,
while they take only 45 minutes to clear dumpers $CD10$ to $CD13$.
Note that clearance time is counted from the moment a group has been completely processed.
In practice, the schedules that the port used previously and which were manually updated by people during the day,
did sometimes not satisfy this constraint.
This indicates that it is a difficult constraint but very important to the people that have to operate the equipment.

\subsection{Stackers}
\begin{figure}[t]
    \centering
    \includegraphics[width=6cm]{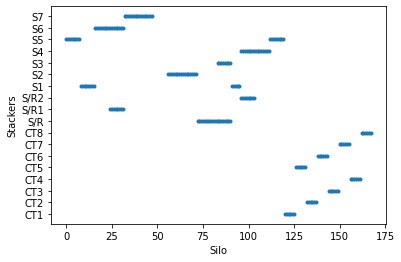}
    \includegraphics[width=6cm]{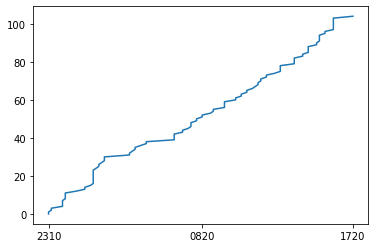}
    \caption{(left) Stacker/silo compatibility. A blue dot means that the stacker can dump onto the stockpile.
    (right) Train arrival time cumulative for one of our instances.}
    \label{fig:stackerSilos}
\end{figure}

Stackers receive bulk material from the conveyors and dump it onto a stockpile.
Stackers also have constraints. The most obvious one is their connection to the conveyor network and the set of stockpiles it has access to.
Indeed, stackers are moving on rail and can access any stockpile in the same row.
Note that in this port there are 19 stackers and there are 168 stockpiles.
Figure~\ref{fig:stackerSilos} (left) shows the compatibility of the stackers and the stockpiles.
As we can see, stockpiles are only reachable by 1 or 2 stackers.
For example S4 and S/R2 share some stockpiles.

\subsection{Equipment flow (path)}
Equipment flows, or paths, are tuples containing one dumper, several conveyors, and one stacker.
The equipment graph contains only 89 feasible paths from dumpers to stackers because of compatibility constraints.
We have observed that the train type C80 has access to 79 paths
and train types C70 and C64 together have access to the remaining 10.
No paths are shared between C80 and the others
and the same set of paths is available for both C70 and C64.
Note that this disparity of paths directly stems from the dumper compatibility while conveyors and stackers are shared between train types.

Furthermore, from the equipment graph, we can extract independent paths 
using any maximum flow algorithm and a capacity of 1 for each piece of equipment.
The objective is to extract the maximum flow going from the dumpers to the stackers.
By doing so, we can see that the maximum flow is 13, which is also the number of dumpers.
The average processing time of a train on CD1 to CD9 is 1h30 and 1h for CD10 to CD13.
Clearance time on CD1 to CD9 is 1h after processing each unit train and 45m for CD10 to CD13 after each group of two unit trains.
This leads us to derive the following naive upper bound $u$ for the total number of unit trains that can be processed in 24 hours: $$u = \frac{24}{2.5} \times 9 + \frac{24}{1.75} \times 4 = 141.3$$. 
This upper bound does not consider the many compatibility constraints and other delays such as are necessary for moving the stackers or trains. 
Clearly, this upper bound is strongly determined by the specified clearance time. In reality, clearance time may actually turn out to be much shorter, but this is not taken into account in the model.

\subsection{Trains}
In addition to the specification of wagon types, cargo types, etc., trains also have an arrival time.
Figure \ref{fig:stackerSilos} (right) shows a cumulative curve of the arrival times of the trains for one day.
As we can see, trains gradually arrive all along the day and night.
Yet it takes hours to process them. Therefore, an efficient model is necessary.

\section{Constraint Programming Model} \label{sec:model}
The previous sections gave a global definition of the constraints and an overview of the specifications of the problem instances.
This section presents the constraint programming model that we use to find the 24 hour schedules.

\subsection{Notation and Parameters}
Let $b \in B$ be a big train and let $t \in T \subseteq B \times \mathbb{N}$ denote a unit train. 
In particular, $(b,i) = t \in T$ refers to $i$-th unit train in a big train.
Let $d \in D$ be a dumper; 
let $s \in STK$ be a stacker; 
let $p \in P$ be a stockpile; 
let $c \in C$ be a conveyor; and
let $l \in EqF$ be a equipment flow.

Additionally,
let $\mathtt{length}_b$ be the length of big train $b$; 
let $\mathtt{load}_t$ be the load of the unit train $t$; 
let $\mathtt{capacity}_p$ be the capacity left on stockpile $p$; 
let $\mathtt{arrival}_t$ be the earliest time that the processing of unit train $t$ can start; 
let $\mathtt{efficient}_{l,t}$ be the efficiency (ton/h) of train $t$ for equipment flow $l$; 
let $\mathtt{clear}_d$ be the dumper $d$ clearance time; and
let $\mathtt{Trans}$ be the transportation time from the shunting yard to the dumpers.

\subsection{Variables}
Table~\ref{tab:model:vars} shows all the variables used by our CP model.
Variables such as $\mathtt{start}_t$ or $\mathtt{end}_t$ are self-explanatory.
The first assumption that we have derived from the instance specifications, is that the path through the equipment graph assigned to a unit train $t$ can be represented as a small integer (namely $\mathtt{path}_t$).
While the number of paths in a graph can growth exponentially
and the equipment graph in our instances has 13 dumpers, 19 stackers, and 62 conveyors,
only 89 paths are feasible once all configuration constraints are enforced.
This can be determined statically.
Moreover, as stated in the specification overview, there are 10 paths for the types C64 and C70,
and 79 paths for the type C80.

The only variable whose meaning is more complicated than the others is $\mathtt{split}_{b,s}$.
The Boolean variable $\mathtt{split}_{b,s}$ indicates that a big train $b$ is split at segment $s$. 
For example, if a big train $b$ with composition $ABC$, then
$\mathtt{split}_{b,0} =$ True indicates that $A$ is separated from $B$ and
$\mathtt{split}_{b,1} =$ True indicates that $B$ is separated from $C$.

\begin{table}[ht]
    \centering
    \begin{tabular}{|l|l|}
        \hline $\forall t \in T$, $\mathtt{start}_t$,$\mathtt{end}_t$,$\mathtt{duration}_t$ &  Train $t$ processing interval.\\
        \hline $\forall t \in T$, $\mathtt{path}_t$ & Equipment path of train $t$.\\
        \hline $\forall b \in B, s \in [0,\mathtt{length}_b-1]$, $\mathtt{split}_{b,s}$ & Is big train $b$ is split at $s$. \\
        \hline $\forall t \in T$, $d \in D$, $\mathtt{dmp}_{t,d}$ & Boolean for dumper use.\\
        \hline $\forall t \in T$, $s \in STK$, $\mathtt{stk}_{t,s}$ & Boolean for stacker use.\\
        \hline $\forall t \in T$, $c \in C$, $\mathtt{cnv}_{t,c}$ & Boolean for conveyor use.\\
        \hline $\forall t \in T$, $p \in P$, $\mathtt{pile}_{t,p}$ & Boolean for stockpile use.\\
        \hline $\forall t \in T$, $\mathtt{alone}_{t}$ & Is train $t$ is alone. \\
        \hline 
    \end{tabular}
    \caption{Variables of the CP model}
    \label{tab:model:vars}
\end{table}

Furthermore, we aggregate some variables for ease of notation:
$\mathtt{Equip}_t$ = $\mathtt{dmp}_{t,1},...,\mathtt{dmp}_{t,|D|},$   $\mathtt{stk}_{t,1},...,\mathtt{stk}_{t,|STK|}, $ $\mathtt{cnv}_{t,1},...,\mathtt{cnv}_{t,|C|}$

\subsection{Constraints}
In this section, we present the mathematical definition of the set of constraints of our model,
after which we further detail each of them independently. 
The complete model is depicted in Figure \ref{fig:cpmodel}.
\begin{figure*}
    \begin{eqnarray}
        \arg \min \ &  \max \limits_{t} (\mathtt{end}_t) \ \text{such that} &         \label{mdl:obj}\\  
       & \mathtt{end}_t = \mathtt{start}_t + \mathtt{duration}_t, &  \forall t \in T  \label{mdl:interval}\\
        &\sum_{t\in T} \mathtt{pile}_{t,p} ~ \mathtt{load}_t \leq \mathtt{capacity}_p, &  \forall p \in P\label{mdl:capacity}\\
        &\text{Table}((\mathtt{path}_t,\mathtt{duration}_t),\mathtt{Eff}_t), &  \forall t \in T \label{mdl:efficiency}\\
        &\text{Table}((\mathtt{path}_t,\mathtt{Equip}_t,\mathtt{alone}_t),\mathtt{Equip\_table}),  & \forall t \in T \label{mdl:equipment}\\
        &\text{Table}((\mathtt{stk}_{t,1},...,\mathtt{stk}_{t,|STK|},
        \mathtt{pile}_{t,1},...,\mathtt{pile}_{t,|P|}),\mathtt{{SP}\_link}),  & \forall t\in T \label{mdl:stkSilo}\\
        &\text{Table}((\mathtt{split}_{b,0},\dots,\mathtt{split}_{b,\mathtt{length}_b-2}, & \nonumber\\
        &\hspace*{1cm}\{\mathtt{dmp}_{t,i},\dots, \mathtt{alone}_t | \forall t \in b\}),\mathtt{Len\_table}), &  \forall b \in B \label{mdl:Tonnage}\\
        &\neg \mathtt{split}_{b,i} \implies \mathtt{start}_{b,i+1} = \mathtt{end}_{b,i} & \forall b \in B, \forall i \in [0, \mathtt{length}_b - 1] \label{mdl:NoSplit}\\
        &\mathtt{arrival}_t +\mathtt{Trans} + \mathtt{STIME} \sum_i \mathtt{split}_{b,i} \leq \mathtt{start}_t & \forall t \in T
        \label{mdl:splitTime} \\
        &\text{unary}([\mathtt{start}_t, \mathtt{end}_t+\mathtt{clear}_d], \mathtt{dmp}_{t,d} \quad \forall t\in T),& \forall d \in D \label{mdl:unaryDMP}\\
        &\text{unary}([\mathtt{start}_t, \mathtt{end}_t], \mathtt{stk}_{t,s} \quad \ \forall t\in T),& \forall s \in STK \label{mdl:unarySTK}\\
        &\text{unary}([\mathtt{start}_t, \mathtt{end}_t], \mathtt{cnv}_{t,c} \quad \ \forall t\in T),& \forall c \in C \label{mdl:unaryCNV}\\
        &\text{unary}([\mathtt{start}_t, \mathtt{end}_t], \mathtt{pile}_{t,p} \quad \ \forall t\in T),& \forall p \in P \label{mdl:unaryPILE}
    \end{eqnarray}
    \caption{Full constraint model}
    \label{fig:cpmodel}
\end{figure*}
First of all, note that all the configuration constraints, 
such as the possible type of material for dumpers etc.,
are directly enforced using the initial domain of the variables.

The objective (\ref{mdl:obj}) limits the makespan by minimizing the maximum ending time.
The constraint (\ref{mdl:interval}) defines the interval relationship between the 
$\mathtt{start}_t$,$\mathtt{end}_t$ and $\mathtt{duration}_t$ variables\footnote{Note that in some solvers, the interval objects are directly accessible.}.
Constraint (\ref{mdl:capacity}) prevents the overflow of stacks.
Constraint (\ref{mdl:NoSplit}) ensures that if no split is performed, 
then the unit trains on the same dumper are sequential require no additional clearance time.
Equation (\ref{mdl:splitTime}) ensures that the minimal splitting time 
plus the transport time from the shunting yard to the dumper is satisfied. We denote the effective splitting time with constant $\mathtt{STIME}$ ($30$ minutes in our problem instances).

\paragraph*{Tables}
Let $\mathtt{\mathtt{Eff}}_t$ be a table containing couples $(path,duration)$ for a given train $t$.
This table can be constructed from the specified equipment flow efficiencies
$\{(l, \mathtt{efficient}_{l,t}, \mathtt{load}_t) | l \in EqF\}$, where $\mathtt{efficient}_{l,t}$ is given as ton/h and $\mathtt{load}_t$ in tons. 
Constraint (\ref{mdl:efficiency}) ensures that the relationship between the selected path
of a train and its total duration is satisfied.

Let $\mathtt{Equip\_table}$ be a table containing tuples of the form:
$(\mathtt{path},\mathtt{dmp}_1,$ \ldots $ ,\mathtt{dmp}_r,\mathtt{stk}_1, $ \ldots $ ,\mathtt{stk}_d,\mathtt{cnv}_1, $ \ldots $ ,\mathtt{cnv}_k,\mathtt{alone})$, in which each of the items except $\mathtt{path}$ is a Boolean.
Each tuple represents a feasible equipment flow.
Note that there only 89 feasible paths, because the dumper type, cargo type, and height all have to agree.
Constraint (\ref{mdl:equipment}) ensures that trains 
can only select feasible equipment flows and that the path number identifies the choice of dumpers, stackers, conveyors, and group length. 

Let $\mathtt{{SP}\_link}$ be a table with Boolean tuples of the form:
$(\mathtt{stk}_1,...,\mathtt{stk}_{|STK|},\mathtt{pile}_1,...,$ $\mathtt{pile}_{|P|})$.
For each compatible couple ($\mathtt{stk}_i,\mathtt{pile}_j$), a tuple with 1's at positions
$i$ and $|STK| + j$ and 0 everywhere else is added.
Constraint (\ref{mdl:stkSilo}) ensures that the compatibility is satisfied and that exactly one stockpile is selected.
Once again, only a few tuples are required to encode the valid combinations of stackers and stockpiles with respect to the current train.

Constraint (\ref{mdl:Tonnage}) handles group processing and length constraints.
More precisely, it ensures that the length of the group and the dumper match.
Splits have to be detected to account for the additional splitting time.
Let $\mathtt{Len\_table}$ be a table as defined in Table~\ref{tab:tonnage}.
In this table, several variables are involved.
First, $s_0$, $s_1$, and $s_2$ correspond to the split variables of a big train. 
For example consider a big train of length 4 defined by $t1-t2-t3-t4$.
$s0$ is the split variable between $t1$ and $t2$.
In Table~\ref{tab:tonnage}, each shade of green corresponds with a length:
\begin{itemize}
    \item Light green is for length 2 with only 1 split variable.
    \item Medium green is for length 3 with 2 split variables.
    \item Dark green is for length 4 with 3 split variables.
\end{itemize}
In addition to the split variables, it also refers to the $\mathtt{dmp}_{t,d}$ variables where $t$ corresponds to a unit train and $d$ to a dumper of length 2.
Finally, the $\mathtt{alone}_{t}$ variables indicate whether the train uses a dumper of length 1.
Notes on the complexity:
Let $T2$ be the set of length 2 dumpers.
The number of tuples in $\mathtt{Len\_table}$ for big trains of length 2 is $|T2|+1$.
Length 2 trains (AB) can be split or not. If not split, the first $|T2|$ tuples are a restriction to the given length 2 dumper. If split (A-B), the final tuple encodes an assignment to any of the length 1 dumpers.
The number of tuples for big trains of length 3 (ABC) is $2\times|T2|+1$.
Basically, the extra unit train doubles the number of cases that were valid for length 2 (A-BC and AB-C),
while the number of cases for length 1 dumpers remains shared. 
Finally, the number of tuples for big trains of length 4 (ABCD) is $|T2|\times(|T2|+1) + 2\times|T2|+1$,
where $|T2|^2$ additional tuples come from a single split in the middle (AB-CD) and the remaining $|T2|$ repeat some of the combinations of length 3 trains (A-B-CD).   
Note that in the instance specifications $|T2| = 4$, which implies that the total number of tuples is
$4 \times 5 + 2 \times 4+1 = 29$.
This is an upper bound as not all the dumpers are valid for a unit train.				

\begin{table}[ht]
\resizebox{\textwidth}{!}{%
\begin{tabular}{ccccccccccccccc}
\multicolumn{3}{c}{Split}                                                                                                                        & T2                                                                             & T2                                                      & \cellcolor[HTML]{FFD966}T1                           & T2                                                      & T2                                                        & \cellcolor[HTML]{FFD966}T1                               & T2                                                        & T2                                                        & \cellcolor[HTML]{FFD966}T1                               & T2                                                        & T2                                                        & \cellcolor[HTML]{FFD966}T1                               \\
\rowcolor[HTML]{E2EFDA} 
\multicolumn{1}{l}{\cellcolor[HTML]{A9D08E}s2} & \multicolumn{1}{l}{\cellcolor[HTML]{C6E0B4}s1} & \multicolumn{1}{l}{\cellcolor[HTML]{E2EFDA}s0} & \multicolumn{1}{l}{\cellcolor[HTML]{E2EFDA}{\color[HTML]{000000} $dmp_{t,i}$}} & \multicolumn{1}{l}{\cellcolor[HTML]{E2EFDA}$dmp_{t,j}$} & \multicolumn{1}{l}{\cellcolor[HTML]{E2EFDA}$\mathtt{alone}_t$} & \multicolumn{1}{l}{\cellcolor[HTML]{E2EFDA}$dmp_{t+1,i}$} & \multicolumn{1}{l}{\cellcolor[HTML]{E2EFDA}$dmp_{t+1,j}$} & \multicolumn{1}{l}{\cellcolor[HTML]{E2EFDA}$\mathtt{alone}_{t+1}$} & \multicolumn{1}{l}{\cellcolor[HTML]{E2EFDA}$dmp_{t+2,i}$} & \multicolumn{1}{l}{\cellcolor[HTML]{E2EFDA}$dmp_{t+2,j}$} & \multicolumn{1}{l}{\cellcolor[HTML]{E2EFDA}$\mathtt{alone}_{t+2}$} & \multicolumn{1}{l}{\cellcolor[HTML]{E2EFDA}$dmp_{t+3,i}$} & \multicolumn{1}{l}{\cellcolor[HTML]{E2EFDA}$dmp_{t+3,j}$} & \multicolumn{1}{l}{\cellcolor[HTML]{E2EFDA}$\mathtt{alone}_{t+3}$} \\
\cellcolor[HTML]{A9D08E}1                      & \cellcolor[HTML]{C6E0B4}1                      & \cellcolor[HTML]{E2EFDA}1                      & \cellcolor[HTML]{E2EFDA}0                                                      & \cellcolor[HTML]{E2EFDA}0                               & \cellcolor[HTML]{E2EFDA}1                            & \cellcolor[HTML]{E2EFDA}0                               & \cellcolor[HTML]{E2EFDA}0                                 & \cellcolor[HTML]{E2EFDA}1                                & \cellcolor[HTML]{C6E0B4}0                                 & \cellcolor[HTML]{C6E0B4}0                                 & \cellcolor[HTML]{C6E0B4}1                                & \cellcolor[HTML]{A9D08E}0                                 & \cellcolor[HTML]{A9D08E}0                                 & \cellcolor[HTML]{A9D08E}1                                \\
\cellcolor[HTML]{A9D08E}1                      & \cellcolor[HTML]{C6E0B4}1                      & \cellcolor[HTML]{E2EFDA}0                      & \cellcolor[HTML]{E2EFDA}1                                                      & \cellcolor[HTML]{E2EFDA}0                               & \cellcolor[HTML]{E2EFDA}0                            & \cellcolor[HTML]{E2EFDA}1                               & \cellcolor[HTML]{E2EFDA}0                                 & \cellcolor[HTML]{E2EFDA}0                                & \cellcolor[HTML]{C6E0B4}0                                 & \cellcolor[HTML]{C6E0B4}0                                 & \cellcolor[HTML]{C6E0B4}1                                & \cellcolor[HTML]{A9D08E}0                                 & \cellcolor[HTML]{A9D08E}0                                 & \cellcolor[HTML]{A9D08E}1                                \\
\cellcolor[HTML]{A9D08E}1                      & \cellcolor[HTML]{C6E0B4}1                      & \cellcolor[HTML]{E2EFDA}0                      & \cellcolor[HTML]{E2EFDA}0                                                      & \cellcolor[HTML]{E2EFDA}1                               & \cellcolor[HTML]{E2EFDA}0                            & \cellcolor[HTML]{E2EFDA}0                               & \cellcolor[HTML]{E2EFDA}1                                 & \cellcolor[HTML]{E2EFDA}0                                & \cellcolor[HTML]{C6E0B4}0                                 & \cellcolor[HTML]{C6E0B4}0                                 & \cellcolor[HTML]{C6E0B4}1                                & \cellcolor[HTML]{A9D08E}0                                 & \cellcolor[HTML]{A9D08E}0                                 & \cellcolor[HTML]{A9D08E}1                                \\
\rowcolor[HTML]{C6E0B4} 
\cellcolor[HTML]{A9D08E}1                      & 0                                              & 1                                              & 0                                                                              & 0                                                       & 1                                                    & 1                                                       & 0                                                         & 0                                                        & 1                                                         & 0                                                         & 0                                                        & \cellcolor[HTML]{A9D08E}0                                 & \cellcolor[HTML]{A9D08E}0                                 & \cellcolor[HTML]{A9D08E}1                                \\
\rowcolor[HTML]{C6E0B4} 
\cellcolor[HTML]{A9D08E}1                      & 0                                              & 1                                              & 0                                                                              & 0                                                       & 1                                                    & 0                                                       & 1                                                         & 0                                                        & 0                                                         & 1                                                         & 0                                                        & \cellcolor[HTML]{A9D08E}0                                 & \cellcolor[HTML]{A9D08E}0                                 & \cellcolor[HTML]{A9D08E}1                                \\
\rowcolor[HTML]{A9D08E} 
0                                              & 1                                              & 1                                              & 0                                                                              & 0                                                       & 1                                                    & 0                                                       & 0                                                         & 1                                                        & 1                                                         & 0                                                         & 0                                                        & 1                                                         & 0                                                         & 0                                                        \\
\rowcolor[HTML]{A9D08E} 
0                                              & 1                                              & 1                                              & 0                                                                              & 0                                                       & 1                                                    & 0                                                       & 0                                                         & 1                                                        & 0                                                         & 1                                                         & 0                                                        & 0                                                         & 1                                                         & 0                                                        \\
\rowcolor[HTML]{A9D08E} 
0                                              & 1                                              & 0                                              & 1                                                                              & 0                                                       & 0                                                    & 1                                                       & 0                                                         & 0                                                        & 1                                                         & 0                                                         & 0                                                        & 1                                                         & 0                                                         & 0                                                        \\
\rowcolor[HTML]{A9D08E} 
0                                              & 1                                              & 0                                              & 0                                                                              & 1                                                       & 0                                                    & 0                                                       & 1                                                         & 0                                                        & 1                                                         & 0                                                         & 0                                                        & 1                                                         & 0                                                         & 0                                                        \\
\rowcolor[HTML]{A9D08E} 
0                                              & 1                                              & 0                                              & 1                                                                              & 0                                                       & 0                                                    & 1                                                       & 0                                                         & 0                                                        & 0                                                         & 1                                                         & 0                                                        & 0                                                         & 1                                                         & 0                                                        \\
\rowcolor[HTML]{A9D08E} 
0                                              & 1                                              & 0                                              & 0                                                                              & 1                                                       & 0                                                    & 0                                                       & 1                                                         & 0                                                        & 0                                                         & 1                                                         & 0                                                        & 0                                                         & 1                                                         & 0                                                       
\end{tabular}}
\caption{Table $T_{T12}$ that contains the constraints with respect to the length. }\label{tab:tonnage}
\end{table}

\paragraph*{Multi-Interval vs. Single-Interval for Unary constraints}
In this section, we discuss two implementations for the unary resource constraints
with \textit{optional activities} (\ref{mdl:unaryDMP})-(\ref{mdl:unaryPILE}).
In the general case, as proposed in \cite{vilim2007global}, each optional activity can be 
rejected or accepted depending on whether the associated intervals necessarily overlap or not. 
In our setting, the average instance contains a 100 trains and almost 300 pieces of equipment,
thus requiring an average total number of 30,000 optional activities.
Optional activities are then linked together using constraints such 
as \textit{minimum} to extract bounds during the search for a given train.

That is why we use another propagation for the unary constraint, which is weaker, 
but requires only 1 interval per train globally.
For the pruning of mandatory task, algorithms are the same.
The difference is in the pruning of \textit{optionals}.
In the classical version, the optional intervals \textit{start}, \textit{end} and \textit{use}
are modified.
In our implementation, the shifted value of \textit{start} is processed,
and if this value is out of bound, then the \textit{use} variable is set to \textit{false}.
We propose here to use the weaker implementation.
In the experimental section, a comparison is given between the two implementations.
Another important point is that the cleaning time of dumpers $CD10$ to $CD13$ 
is only added to the second train $end$ variable, and not to both.

\subsection{Search Strategies}
We propose here to consider the \textit{PathThenStart} search strategy.
This strategy works in two phases:
First, the strategy selects the next path by maintaining information for each dumper
(i.e. the minimal finishing time of the already scheduled trains is used to select for each train an equipment flow).
Then, once all the equipment flows have been assigned, the search starts assigning the other variables ($\mathtt{start}_t$, etc.)

The main reason why we choose to use this search is because trains are arriving in a given order. Then, the second part of the search might be easy to solve heuristically, and not necessarily useful to solve optimally.
Then the search can restart and look for a better assignment of the equipment flows.
This search, combined with large neighborhood search is providing the best results with this model in our experiments.

\section{Results} \label{sec:results}
In this section, we analyse the results of the model proposed in this paper.
We start by giving the method used to generate a large set of instances with data 
fitting the statistical distribution of the real instances.
Then, we first show an example of result on a real data set, 
before providing a large panel of results on the generated data sets.

\begin{figure}[b!t]
    \centering
    \includegraphics[width=9cm]{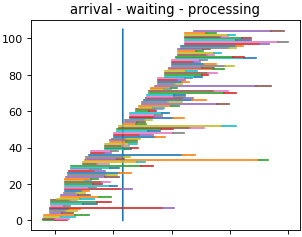}
    \caption{Solution for a real instance. For each row (i.e. train), the first colored segment define the waiting time, the second colored segment the processing time. The $x$ axis is the time, the value on the $y$ axis is the train id.}
    \label{fig:solution}
\end{figure}
\begin{figure}[h!t]
    \centering
    \includegraphics[width=12cm]{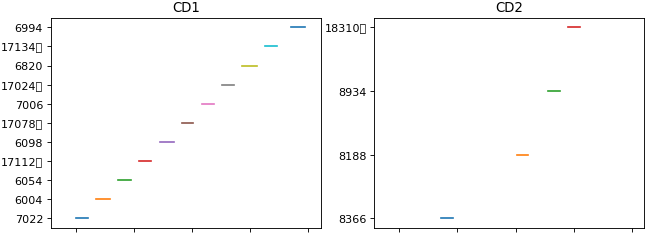}
    \caption{Usage of dumpers CD1 and CD2. Each segment is a train, the value on the $y$ axis is the train name. The $x$ axis is the time.}
    \label{fig:solution2}
\end{figure}
\subsection{Instance generation}
We generated a large number of instances by varying the number of big trains.
In each of these instances the port remains the same.
More precisely, for each number of big train $|O| \in [5, 10, 15, 17, 20, 22, 25, 30, 35, 40, 45, 50, 55, 60]$, we generated 30 instances.
Each big train contains between 1 and 4 unit trains.
The number of unit trains, the type of material, the quantity, the time between the arrival of two big trains etc., all this is randomly generated from an analysis of the real instances.
Note that in our real instances, the number of train varies from 80 to 110,
in the generated data, it varies from 7 to 166.
This section presents cumulative results on the 420 generated instances.
The run was limited to 3 minutes and no parallelism was allowed.

{\em 1. Real instance}. ~
The real instance solved in Figure~\ref{fig:solution} shows the \textit{arrival-waiting-processing} operation where each segment is made of two colors.
The first one is the arrival and waiting time, the second one represents the processing time.
As we can see in Figure \ref{fig:solution2}, some dumpers like CD1 are heavily used, while some others are not.

\begin{figure}[h!]
    \centering
    \includegraphics[width=12cm]{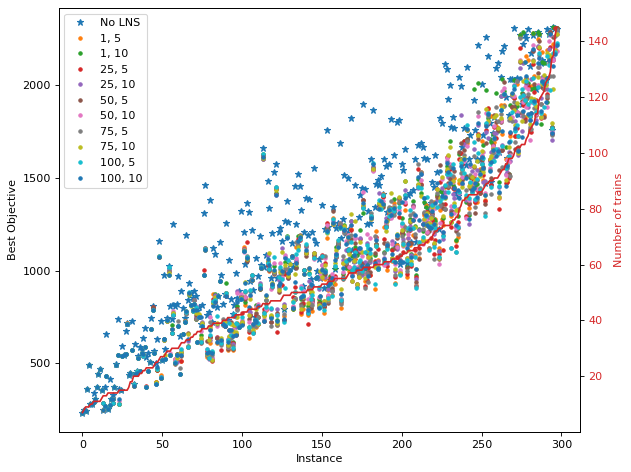}
    \caption{No LNS vs LNS. 
    The first number is the factor of the Luby restart policy, 
    the second is the length of the destroy.
    The red curve is the number of trains.}
    \label{fig:LNS1}
\end{figure}

{\em 2. Generated instances}. ~
The generated instances are solved and the results plotted in Figure~\ref{fig:LNS1} with the \textit{No LNS} points.
The $Y$ axe represents the obtained makespans.
The $X$ represents the index of the instances, sorted by number of train in the instance.
The \textit{No LNS} method is the \textit{PathThenStart} search heuristic up with the model in our internal solver.
The table algorithm are either from \cite{demeulenaere2016compact}, 
or our dedicated version for less than 64 tuples.

\subsection{Multi-Interval VS Single-Interval}
\begin{figure}[ht!]
    \centering
    \includegraphics[width=12cm]{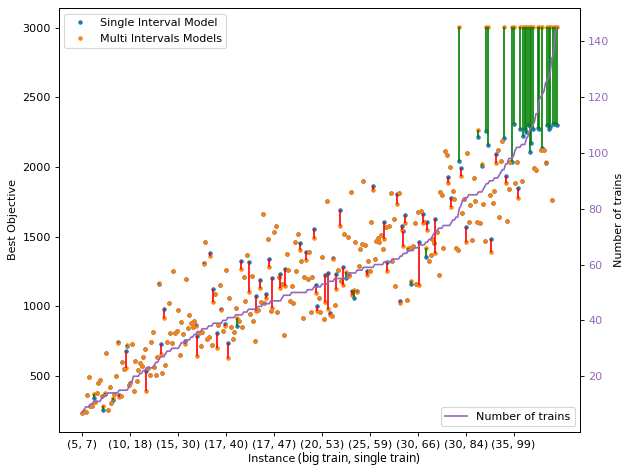}
    \caption{Single-interval model VS Multi-Intervals Model.}
    \label{fig:singleMulti}
\end{figure}

Figure~\ref{fig:singleMulti} shows the impact of the unary implementation for the performances.
In this plot, the green vertical segments show the improvement of the single interval compared to multi interval.
The red vertical segments show the contrary.
In addition, the purple curve shows the number of trains in the instance.
As we can see, when the number of train is small, the multi-interval
is clearly dominating.
Indeed the method allows a finer-grain pruning.
But as the number of trains grows, the single-interval method starts
to be the only one able to solve the problem.
When we profiled the runs of the multi interval, 
the time was spent updating the many optional intervals.
There are around 30 instances solved by the single interval 
that are not solved by the multi-interval.
There is no instance solved by the multi-interval that is not solved by the single-interval.
The transition is around 90 trains, which is slightly above the lower bound
of the sizes of our real instances. 
That is the reason why we use the single interval in practice.

\subsection{Large Neighbours Search (LNS)}
In order to improve the results, and since our search is static with respect to restart, 
we propose to use LNS \cite{pisinger2019large,schaus2011solving}.
Our LNS is simple, we put in place a Luby restart policy, where the factor for the luby is a parameter.
Then we destroy $K$ consecutive train schedules, where $K$ is also a parameter and we reschedule them at random.
Figure~\ref{fig:LNS1} shows the results of our LNS.
LNS $(i, k)$ means $i$ times the luby criteria before restarting, and at each destroy, $k$ consecutive trains are unscheduled.
In the left plot, we sorted the instances with respect to the best objective value of the \textit{No LNS} method.
As we can see, the \textit{No LNS} is completely dominated by all the LNS methods.
Incorporating the LNS in our model was a strict improvement in the solution quality.
We tried larger value for the destroy (up to 20), but no LNS configuration was dominating the others.
The configuration that empirically gave the best result is \textit{(25,10)}.
Finally, Figure~\ref{fig:LNS1} shows that the number of trains 
in the instance changes the order of performances of the LNS.
Such results may imply that the parameters of the LNS should be a function of the number of trains.

\section{Conclusion}
In this paper, we proposed a new train-scheduling model using constraint-programming for the scheduling community
named \sump.
The proposed model was able to solve the problem up to 100 trains, which was the industrial objective.
We proposed a simple yet efficient LNS to improve the results.
Then, we generated a large set of generated instances ranging from 7 to 166 trains.
The goal is that this set of instances will help the development of future
scheduling models and propagators for this type of problem, as shown in this paper with
the different implementation of sub-parts of the model, 
and the different implementations of propagators.

\bibliography{ts-ref}
\bibliographystyle{plain}

\end{document}